\pdfoutput=1

\documentclass[11pt]{article}

\usepackage[]{emnlp2021}

\usepackage{times}
\usepackage{latexsym}

\usepackage[T1]{fontenc}

\usepackage[utf8]{inputenc}

\usepackage{microtype}

\usepackage{graphicx}
\usepackage{amssymb}

\title{Logic-level Evidence Retrieval and Graph-based Verification Network for Table-based Fact Verification}

\author{Qi Shi,\ \ Yu Zhang\thanks{\ \ Corresponding author.},\ \ Qingyu Yin,\ \ Ting Liu \\
 \normalsize{Research Center for Social Computing and Information Retrieval}\\[-.05cm]
 \normalsize{Harbin Institute of Technology, Harbin, China}\\[-.05cm]
 {\small\tt \{qshi,zhangyu,qyyin,tliu\}@ir.hit.edu.cn}  \\
}

\begin{document}
\maketitle
\begin{abstract}
Table-based fact verification task aims to verify whether the given statement is supported by the given semi-structured table.
Symbolic reasoning with logical operations plays a crucial role in this task. 
Existing methods leverage programs that contain rich logical information to enhance the verification process. 
However, due to the lack of fully supervised signals in the program generation process, spurious programs can be derived and employed, which leads to the inability of the model to catch helpful logical operations.
To address the aforementioned problems, in this work, we formulate the table-based fact verification task as an evidence retrieval and reasoning framework, proposing the \textbf{L}ogic-level \textbf{E}vidence \textbf{R}etrieval and \textbf{G}raph-based \textbf{V}erification network (LERGV). Specifically, we first retrieve logic-level program-like evidence from the given table and statement as supplementary evidence for the table.
After that, we construct a logic-level graph to capture the logical relations between entities and functions in the retrieved evidence, and design a graph-based verification network to perform logic-level graph-based reasoning based on the constructed graph to classify the final entailment relation.
Experimental results on the large-scale benchmark TABFACT show the effectiveness of the proposed approach\footnote{Our code is available at: \url{https://github.com/qshi95/LERGV}}.
\end{abstract}

\section{Introduction}

There are a large number of semi-structured tables on the Internet. How to perform reasoning over semi-structured tables is crucial for people to understand different types of information in the real world. And this direction has spawned many tasks. Among these tasks, table-based fact verification task has recently received a lot of attention, which is important for many applications, such as fake news detection, scientific paper understanding \cite{wang2021semeval}, etc. This task aims to verify the correctness of the given statement by the given table, which requires both linguistic reasoning and symbolic reasoning \cite{chen2019tabfact}.

\begin{figure}
    \centering
    \includegraphics[width=0.45\textwidth]{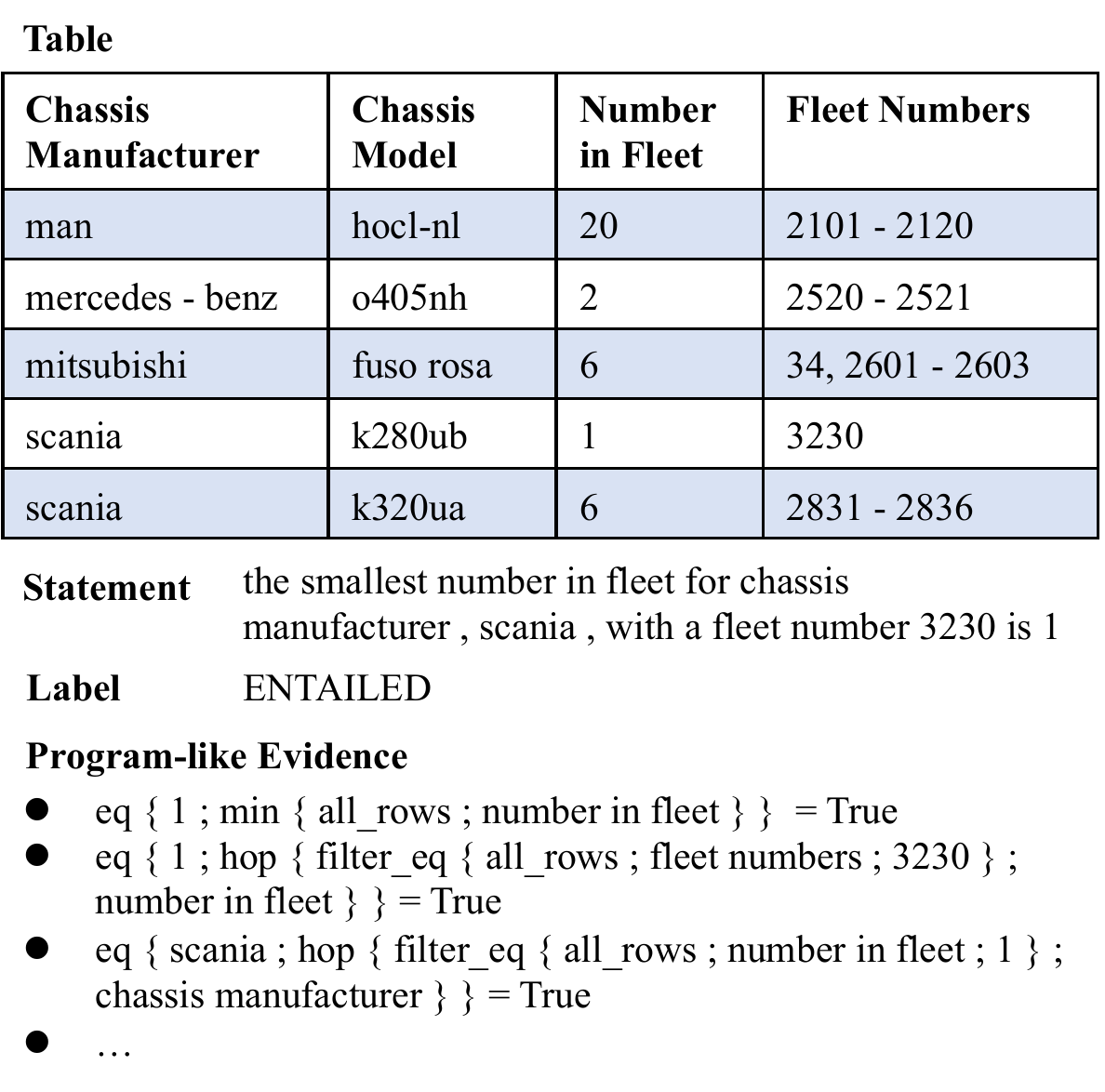}
    \caption{Example of TABFACT dataset. Given a table and a statement, the goal is to verify the correctness of the statement by the table. 
    Program-like evidence contains logical operations and can be used as supplementary information to a table. } 
    \label{example} 
\end{figure}

Symbolic reasoning with logical operations like \emph{"count"} and \emph{"argmax"} plays an important role in the table-based fact verification task. Figure \ref{example} shows an example.
Ideally, to verify the correctness of such statements, logical operations provide strong hints to classify the entailment relation.
Therefore, how to utilize such logical operations is crucial in this task.



Program is a kind of logic form derived from tables, which contains rich logical operations. Following \cite{chen2019tabfact}, existing methods \cite{zhong2020logicalfactchecker,yang2020program,shi2020learn} mostly use programs to perform symbolic reasoning. Specifically, they derive one or several programs with a semantic parser based on the given table and statement, and then leverage the obtained programs to enhance the verification process. 
However, these models may select spurious programs (i.e. wrong programs with correct returned labels) because there are weak supervised signals in the semantic parsing process.
Consequently, label-consistent programs (the programs whose execution results are consistent with the ground-truth verification labels) tend to be selected instead of semantic-consistent ones. As a result, the obtained programs will not contain helpful logical operations, thus cannot benefit the verification of the correctness of the statement.
Ideally, a natural way of leveraging programs is to regard them as supplementary evidence for tables. In other words, some of the programs contain the necessary information with logical operations that summarize or describe the facts observed from the table. So that we can leverage the information conveyed by the programs to help better capture the facts of the table and then classify the entailment relation. 

Based on the above considerations, in this work, we formulate table-based fact verification as an evidence retrieval and reasoning pipeline, proposing a logic-level evidence retrieval and graph-based verification method, named LERGV.
Firstly, instead of deriving label-consistent programs with a semantic parser, we propose a rule-based method to retrieve valuable logic-level program-like evidence from the table and statement to avoid the issue of spurious programs.
Then we leverage the structure of the programs to construct a logic-level graph with the above evidence to catch the logical relations between entities (such as \emph{"number in fleet"}) and functions (such as \emph{"min"}) in the programs. 
Finally, a graph-based verification network is proposed to reason over the constructed graph to perform logic-level reasoning, which takes advantage of the combination of linguistic reasoning and symbolic reasoning to make the final prediction.

We conduct experiments on a large-scale benchmark dataset TABFACT \cite{chen2019tabfact}.
Experimental results show that our model surpasses all baseline systems with a considerable margin.
The main contributions of this paper are three-fold:
\begin{itemize}
    \item We formulate the table-based fact verification task as an evidence retrieval and graph-based reasoning framework by regarding programs as additional evidence instead of building a weakly supervised semantic parser to avoid the issue of spurious programs.
    \item We construct a logic-level graph and propose a graph-based verification network to catch logical relations between entities and functions in evidence, which can take advantage of both linguistic reasoning and symbolic reasoning.
    \item Experimental results on the TABFACT show the effectiveness of our proposed approach that our method outperforms all baseline systems and achieves competitive results.
\end{itemize}

\begin{figure}
    \centering
    \includegraphics[width=0.45\textwidth]{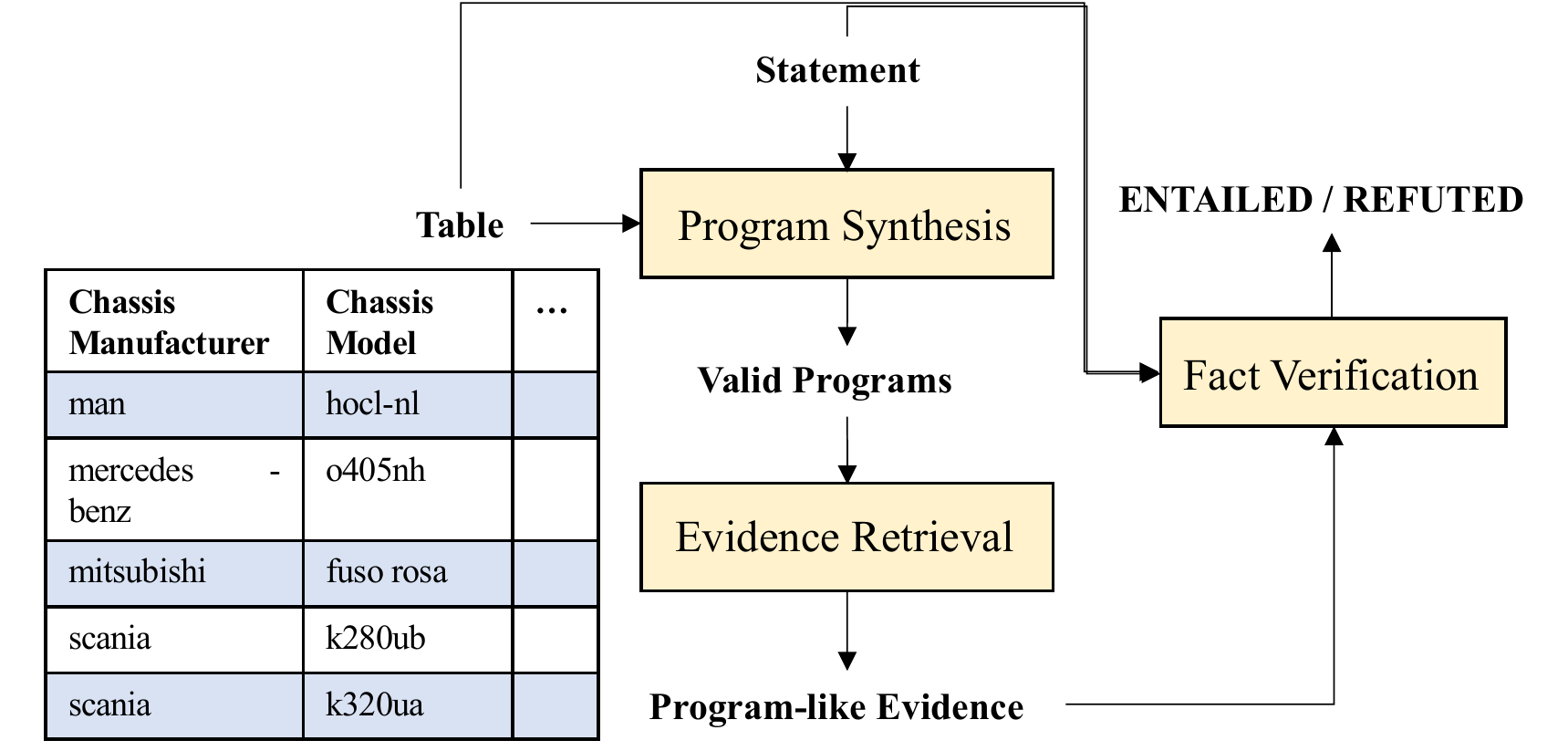}
    \caption{Pipeline for table-based fact verification task. Given a table and a statement, the proposed pipeline can be divided into program synthesis module, evidence retrieval module ($\S \ref{er}$), and fact verification module ($\S \ref{gci}$ and $\S \ref{llgn}$).}
    \label{framework}
\end{figure}

\section{Task Definition and Overview}
\label{td}
In this paper, we study the task of table-based fact verification. Given a table $T$ with $R$ rows and $C$ columns and a statement $S$, the goal is to verify the correctness of the given statement by the given table with the label ENTAILED or REFUTED.

\begin{figure*}
    \centering
    \includegraphics[width=0.9\textwidth]{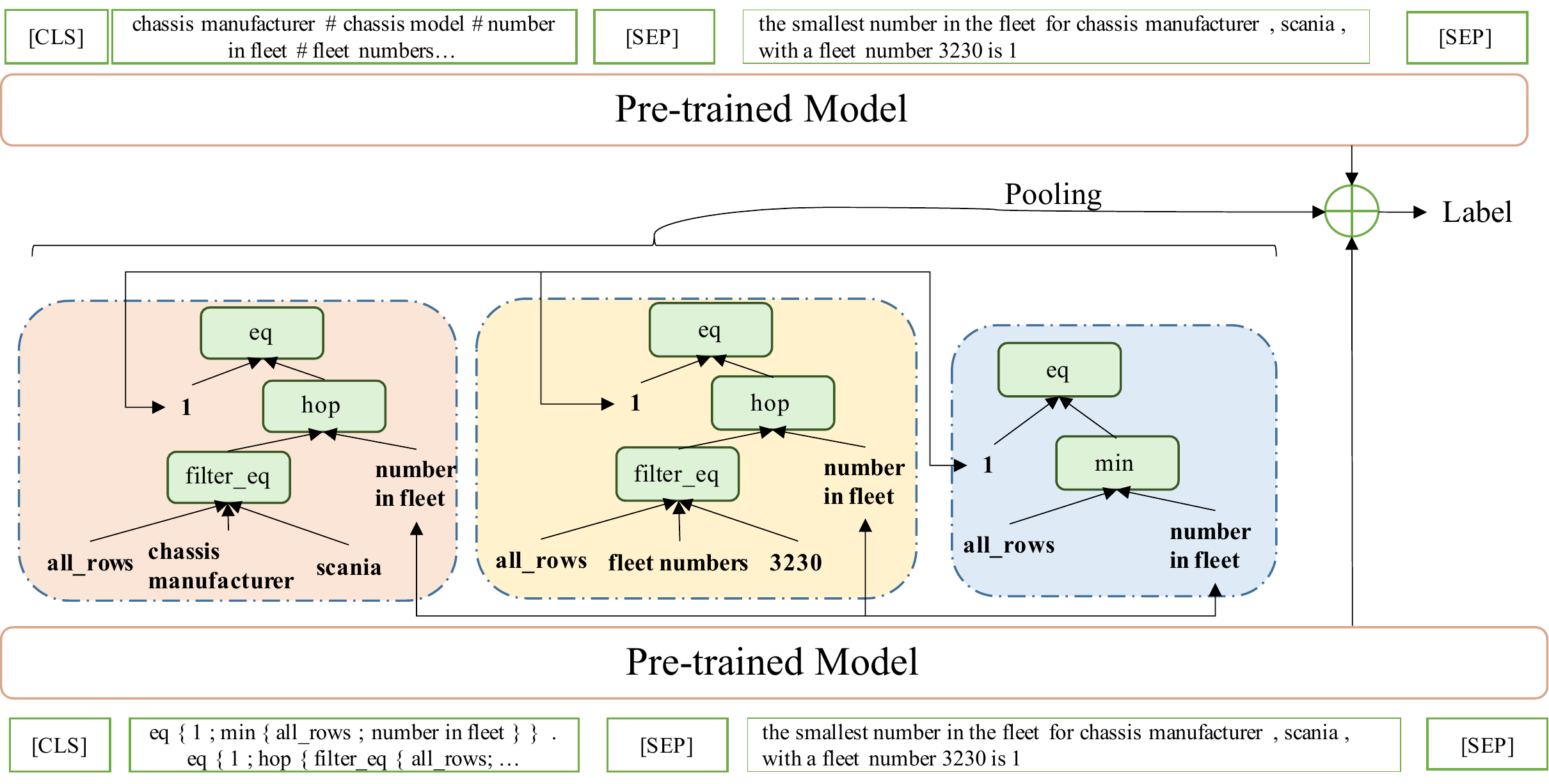}
    \caption{Model structure of LERGV. Logic-level program-like evidence is obtained from the table and statement described in $\S \ref{er}$. Take the table-statement pair and evidence-statement pair as input, we first construct a logic-level graph to catch the logical relations between entities and functions in the programs ($\S \ref{gci}$). After that, we design a graph-based verification network to perform reasoning over the constructed graph before making the final verification ($\S \ref{llgn}$).}
    \label{model}
\end{figure*}

Table is the only evidence in the original task setting. However, we believe that the additional evidence that contains logical operations is helpful to classify the entailment relation. In this study, we employ such evidence (denote as "evidence" in the rest of the paper) that in the form of programs. In particular, program is a kind of LISP-like logical form that follows a grammar with over 50 pre-defined functions \cite{chen2019tabfact}. Every program is tree-structured, consisting of functions as parent nodes and their arguments as children nodes, where leaf nodes represent arguments, namely entities linked to the table or the statement, and non-leaf nodes represent functions, such as \emph{"min"}, \emph{"count"}, \emph{"argmax"}, etc. The dotted boxes in Figure \ref{model} show the structure of programs.

To better take advantage of program-like evidence, in this work, we formulate the table-based fact verification task as an evidence retrieval and reasoning pipeline that consists of three main components, program synthesis, evidence retrieval, and fact verification modules.
Figure \ref{framework} shows the overview of our proposed approach. Given the table and the statement, the program synthesis module first synthesizes all possible programs with valid combinations. Then the evidence retrieval module selects, decomposes, and filters over the programs to obtain valuable logic-level evidence as supplementary information to the original table. Finally, a fact verification model takes the input of the table, statement, and obtained evidence to predict whether the statement is supported by the given table.
Following previous work \cite{zhong2020logicalfactchecker,yang2020program,shi2020learn}, we perform program synthesis with the latent program search algorithm (LPA) \cite{chen2019tabfact}, followed by an evidence retrieval module and a fact verification module.
We will present above modules in $\S$ \ref{section:method}. 

\section{Methodology}
\label{section:method}
We propose a \textbf{L}ogic-level \textbf{E}vidence \textbf{R}etrieval and \textbf{G}raph-based \textbf{V}erification network (LERGV) for the table-based fact verification task.
Given a table and a statement, LERGV works as follows. 
First, we start with the latent program algorithm \cite{chen2019tabfact} to synthesize programs and then use a rule-based retrieval approach to select, decompose, and filter among all synthesized programs to obtain valuable logic-level evidence ($\S$ \ref{er}).
After that, we construct a graph based on the retrieved evidence and initialize graph node representations from a pre-trained language model ($\S$ \ref{gci}).
Finally, we propose a graph-based verification network centered around the obtained evidence to perform graph-based reasoning to predict the final verification result ($\S$ \ref{llgn}).


\subsection{Logic-level Evidence Retrieval}
\label{er}
Program is a kind of logic form with rich logical operations. For the specific task of table-based fact verification, we believe that the program-like evidence can provide valuable information in addition to tables. 
Following \cite{zhong2020logicalfactchecker,shi2020learn,yang2020program}, in this work, we follow the latent program search algorithm (LPA) \cite{chen2019tabfact} to synthesize valid programs with pre-defined functions. 
Given a table $T$ and a statement $S$, LPA first performs entity linking to detect all the entities in the statement and link them to the table, then collects a set of programs by executing sub-programs over the table and store the generated intermediate variables recursively. 

After obtaining the program set $\mathbb{P}=\{(P_i, A_i)\}_{i=1}^N$ for a given statement $S$ (where $P_i$ stands for the i-th program, and $A_i$ refers to the corresponding returned label executing over the table, namely $True$ or $False$), instead of building a semantic parser, we select, decompose, and filter some of them by a series of rules to keep higher quality programs as evidence due to the limitation of input size of the pre-trained language model. Specifically, we retrieve the evidence in the following steps:

\begin{itemize}
    \item We choose programs with the returned label $A_i=True$ in the program set $P$ as evidence to ensure that the evidence can be correctly observed from the table.
    \item We decompose the evidence containing function \emph{"and"} into two separate pieces of evidence, where each one is a subtree of the \emph{"and"} node, to simplify and remove duplicate evidence. Since the label of the original evidence $A_i=True$, the two programs connected by the \emph{"and"} node must both be $True$, thus we guarantee the correctness of the obtained evidence.
    \item For the cases that obtained a large number of programs, we remove evidence that contains functions with negative meanings, including \emph{"not\_eq"}, \emph{"filter\_not\_eq"}, \emph{"not\_within"}, etc. This is due to that programs with negative functions tend to be descriptions of the statements that don’t explicitly exist in tables (e.g. "Number of teams is not 3"), which are less effective than programs with positive functions. This operation can limit the number of evidence and obtain semantically more relevant evidence.
\end{itemize}

So far, by aggregating, decomposing, and filtering the information contained in the table, we can use the logic-level evidence to supplement the original table, thereby enhancing the ability of our model to understand semi-structured tables. Just like the motivating example shown in the Figure \ref{example}, with the evidence containing \emph{"min"}, \emph{"number in fleet"} and \emph{"1"}, model can easily establish connections between \emph{"smallest"}, \emph{"number in fleet"} and \emph{"1"}, which benefits to the verification of the correctness of the statement. 
After obtaining the evidence set $E$, we use program-like evidence directly instead of converting programs into text \cite{yang2020program} to preserve the logical connections between entities and functions in the program. 

\subsection{Graph Construction and Initialization}
\label{gci}
Given the retrieved evidence, we construct a graph to capture the logical relations among all entities and functions in the evidence programs. Specifically, we denote a graph as $G=\{V,E\}$ and treat each function (such as \emph{"filter\_eq"}, \emph{"hop"}) and entity (such as \emph{"all\_rows"}, \emph{"chassis manufacturer"}) as a graph node. Besides, to distinguish between nodes of different origins, we further divide the nodes into two types, namely function nodes and entity nodes (shown as nodes with/without frame in Figure \ref{model}). 

We leverage the structure of evidence to construct edges. In this way, we can explore the connections between entity nodes and function nodes, which will benefit the following verification process.
Specifically, we retain the structure of every program by adding edges between each entity node and its corresponding function node to learn the semantic compositionality of the program. 
Besides, we add edges between entity nodes with the same content across the entire evidence set $E$ to turn the graph into a connected graph to reason over multiple evidence programs, which are shown as arrows in Figure \ref{model}.

We feed the table-statement pair and evidence-statement pair into a pre-trained language model separately as shown in Figure \ref{model} and initialize node representations from the top-layer output of the pre-trained language model. For the node with multiple word pieces, we perform average pooling over the representations of their corresponding positions.

\begin{figure*}
    \centering
    \includegraphics[width=0.9\textwidth]{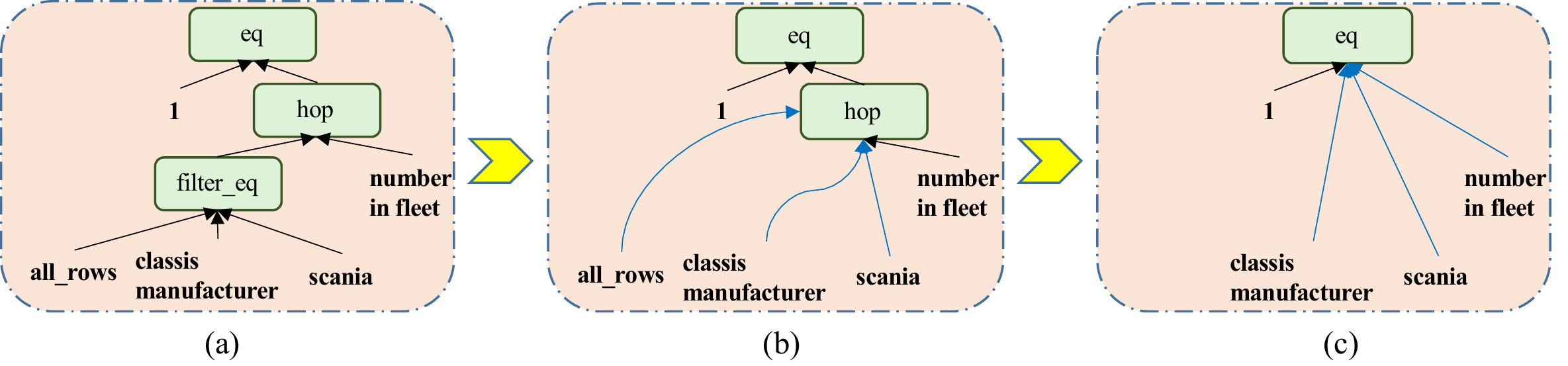}
    \caption{Example of the node pruning process. We take one single program as an example. We remove \emph{"filter\_eq"} in Figure (a) and add edges between its children nodes and parent node. After that, we remove nodes \emph{"all\_rows"} and \emph{"hop"} in the Figure (b), which is the same as Figure (a).}
    \label{node-pruning}
\end{figure*}

\subsection{Graph-based Verification Network}
\label{llgn}
We propose a graph-based verification network, which is designed to perform logic-level reasoning over the retrieved evidence along with the given table and statement to combine linguistic reasoning and symbolic reasoning to benefit the final verification decision.

\paragraph{Graph-based Reasoning Process}
After the graph construction and node initialization, LERGV performs reasoning over the constructed graph. Our network is designed based on the graph attention network \cite{velivckovic2017graph}, to learn the importance between different nodes and fuse the neighbors to perform graph-based reasoning.
Specifically, the node representations are updated as follows:
\begin{equation}
    e_{i,j} = \mathrm{LeakyReLU}(a[W_qh_i||W_kh_j])
\end{equation}
\begin{equation}
    \alpha_{i,j} = \frac{\mathrm{exp}(e_{i,j})}{\sum \limits_{k \in N_{i}} \mathrm{exp}(e_{i,k})}
\end{equation}
\begin{equation}
\label{res}
    h_i=\sigma(\sum \limits_{j \in N_i} \alpha_{i,j} \mathrm{W_v} h_j) + h_i
\end{equation}

where $W_q \in \mathbb{R}^{F \times F}$, $W_k \in \mathbb{R}^{F \times F}$, $W_v \in \mathbb{R}^{F \times F}$, $a : \mathbb{R}^F \times \mathbb{R}^F \rightarrow \mathbb{R}$ are trainable parameters. $N_i$ presents the neighbors of node $i$. $||$ stands for concatenation operation. $h_i$ and $h_j$ mean the representations of node $i$ and node $j$. We will describe them in details in the following parts.

\paragraph{Node Type Representations}
Notice that the nodes are either entities linked to the table or the statement, or functions pre-defined in LPA, to this end, we model different types of nodes in the message passing process. First, we combine type embedding for every node in the graph as follows:
\begin{equation}
    h_i^t = W_t(o_i) + b_t
\end{equation}
\begin{equation}
    h_i = [h_i^p || h^s || h_i^t]
\end{equation}

where $o_i \in \mathbb{R}^T$ is a one-hot vector indicating the type of the node $i$. $W_t \in \mathbb{R}^T \times \mathbb{R}^F$, $b_t \in \mathbb{R}^F$ are trainable parameters. $T$ presents the number of node types, in this work, $T$ is set to $2$. $h_i^p$ means node representation obtained by the node initialization process. 
And $h^s$ means the representation of the statement, which is obtained similarly as $h_i^p$.

\paragraph{Node Pruning}
Programs are designed for executing over the table $T$ recursively. However, as an evidence program, many nodes are semantically unrelated to the statement, such as \emph{"filter\_eq"}, \emph{"hop"}, \emph{"all\_rows"}, etc. Here we propose a node pruning approach to prune the nodes automatically to filter such nodes. Figure \ref{node-pruning} shows an example. We first calculate the relevance score between the node and statement as follows:
\begin{equation}
    s_i = \sigma (W_s[h_i^p || h^s] + b_s)
\end{equation}

where $W_s \in \mathbb{R}^{2F \times F}$, $b_s \in \mathbb{R}^F$ are trainable parameters. 
Then, we remove the nodes with the lowest scores with the probability $\theta$. Finally, take one removed node as an example, we add edges between its children nodes and parent node to connect the new graph to perform graph-based verification.

\paragraph{Label Prediction}
We adopt an attentive pooling layer to obtain the final representation $h$. 
Then, we concatenate $h$ and two [CLS] tokens, that come from the output of the pre-trained language model with table-statement pair and evidence-statement pair as input respectively. 
Finally, we feed the obtained vector into a classifier to predict the probability of each label.
The concatenation operation aims to combine linguistic information containing in the table-statement pair and logic-level information containing in the program-like evidence together to achieve the goal to combine linguistic reasoning and symbolic reasoning.

\begin{table*}[t]
    \centering
    \begin{tabular}{p{6.5cm}ccccc}
    \hline
        Model & Val & Test & Test (simple) & Test (complex) & Small Test  \\
        \hline
        BERT classifier w/o Table & 50.9 & 50.5 & 51.0 & 50.1 & 50.4 \\
        \hline
        Table-BERT-Horizontal-F+T-Concatenate & 50.7 & 50.4 & 50.8 & 50.0 & 50.3 \\
        Table-BERT-Vertical-F+T-Template & 56.7 & 56.2 & 59.8 & 55.0 & 56.2 \\
        Table-BERT-Vertical-T+F-Template & 56.7 & 57.0 & 60.6 & 54.3 & 55.5 \\
        Table-BERT-Horizontal-F+T-Template & 66.0 & 65.1 & 79.0 & 58.1 & 67.9 \\
        Table-BERT-Horitonzal-T+F-Template & 66.1 & 65.1 & 79.1 & 58.2 & 68.1 \\
        \hline
        LPA-Voting w/o Discriminator & 57.7 & 58.2 & 68.5 & 53.2 & 61.5 \\
        LPA-Weighted-Voting & 62.5 & 63.1 & 74.6 & 57.3 & 66.8 \\
        LPA-Ranking w/ Transformer & 65.2 & 65.0 & 78.4 & 58.5 & 68.6 \\
        \hline
        LogicalFactChecker & 71.8 & 71.7 & 85.4 & 65.1 & 74.3 \\
        HeterTFV & 72.5 & 72.3 & 85.9 & 65.7 & 74.2 \\
        SAT & 73.3 & 73.2 & 85.5 & 67.2 & - \\
        ProgVGAT & 74.9 & 74.4 & \bf 88.3 & 67.6 & 76.2 \\
        \hline
        LERGV & \bf 75.6 & \bf 75.5 & 87.9 & \bf 69.5 & \bf 77.8 \\
        Human Performance & - & - & - & - & 92.1 \\
        \hline
        
    \end{tabular}
    \caption{Experimental results on TABFACT. For Table-BERT, \emph{T} and \emph{F} refer to table and statement respectively. \emph{Horizontal} and \emph{Vertical} represent the scanning strategies of linearizing tables. \emph{Concatenate} and \emph{Template} stand for whether use templates to concatenate the table cells. For LPA, \emph{Voting} and \emph{Weighted-Voting} mean voting for the result without/with a weighted-sum score. \emph{Ranking} means using the result of the top-ranked program. 
    }
    \label{tab:exp_result}
\end{table*}

\section{Experiments}
\subsection{Dataset and Experimental Settings}
We evaluate our model on the TABFACT \cite{chen2019tabfact}, a large-scale dataset for table-based fact verification, which contains $92283$, $12792$, and $12779$ samples in training, validation, and test sets respectively, with one table and one statement in each sample. Each sample is labeled as either ENTAILED or REFUTED, indicates whether the statement is supported by the table. The test set is further divided into the simple channel and complex channel to distinguish the difficulty, with $4171$ and $8608$ samples for each. Besides, a small test set with $2K$ samples is provided for human evaluation. Samples in the dataset require symbolic reasoning with logical operations, such as \emph{"min"}, \emph{"argmax"}, \emph{"count"}, etc. Following the existing work, we use accuracy as the evaluation metric.

Following \cite{chen2019tabfact}, we use BERT-base \cite{devlin2019bert} as the backbone to build our model.
The maximum sequence length is $512$, the batch size is set to $8$, the learning rate is set to $1e-5$, the warmup step is set to $3000$, and the probability of pruning nodes $\theta$ is set to $0.3$. We set the size of all hidden layers to $768$, which is the same as the output of the BERT-base model. Cross entropy loss is adopted to optimize the model.

\subsection{Baseline Systems}
\paragraph{Latent Program Algorithm (LPA)}
LPA \cite{chen2019tabfact} treats the task in a weakly supervised manner by feature-based entity linking, latent program generation, and candidate program ranking with a Transformer-based encoder \cite{vaswani2017attention}.
\paragraph{Table-BERT}
Table-BERT \cite{chen2019tabfact} views the task as a semantic matching problem by encoding linearized table and the statement via BERT to predict the final label.
\paragraph{LogicalFactChecker}
LogicalFactChecker \cite{zhong2020logicalfactchecker} derives one program with different semantic parsers and represents it with graph module networks to learn the semantic compositionality of the program.
\paragraph{HeterTFV}
HeterTFV \cite{shi2020learn} chooses multiple latent programs and proposes a heterogeneous graph-based reasoning network to reason over different types of information.
\paragraph{SAT}
SAT \cite{zhang2020table} proposes a structure-aware table representation method by utilizing the mask in self-attention layers.
\paragraph{ProgVGAT}
ProgVGAT \cite{yang2020program} improves the semantic parser with a specific loss function and converts the obtained program to natural language sentences with pre-defined templates to perform graph-based reasoning.

\subsection{Experimental Results}
Table \ref{tab:exp_result} shows the experiment results: our model reaches an accuracy of $75.5\%$ on the test set, which surpasses all baseline systems with remarkable improvements. \footnote{We do not compare with the TAPAS-based approach directly \cite{eisenschlos2020understanding}. Because our approach is centered around evidence rather than table. The pre-trained model with evidence-statement pair as input is not adapted to the TAPAS model.}

From Table \ref{tab:exp_result}, we can observe that our model outperforms LPA \cite{chen2019tabfact}, Table-BERT \cite{chen2019tabfact}, and SAT \cite{zhang2020table} with large margins, which illustrates the advantage of the combination of linguistic reasoning and symbolic reasoning.
Meanwhile, compared with the semantic parsing-based methods, i.e., LogicalFactChecker \cite{zhong2020logicalfactchecker}, HeterTFV \cite{shi2020learn}, and ProgVGAT \cite{yang2020program}, our model gains improvements in performance from $1.1\%$ to $3.8\%$, which indicates the effectiveness of our proposed model that our model can better understand the semi-structured tables by retrieving logic-level evidence as supplementary information and catching logical relations between entities and functions.
In addition, we can also see that our model surpasses ProgVGAT \cite{yang2020program} by nearly $2$ points on the complex test set, which further shows the ability of the proposed graph-based verification network on dealing with complex statements. 
In sum, all these results demonstrate the utility of the proposed method for fact verification over semi-structured tables. 

Notice that there is a narrow gap between our model and ProgVGAT \cite{yang2020program} on simple test set, which is due to that our method retrieves related programs with rich logical operations, which naturally brings more benefits to complex statements. In comparison, ProgVGAT \cite{yang2020program} focuses on building a better semantic parser and only choose the most suitable program, so that works better on simple statements. Moreover, our method also gains competitive performance on the simple testset, with only a 0.4\% gap compared with ProgVGAT \cite{yang2020program}.

\subsection{Ablation Study}


We report how each component contributes to LERGV by eliminating each one from the entire model on the validation set. Table \ref{tab:ablation} shows the results. 
Specifically, the removals of the node pruning module, node type representation module lead to a drop by $0.4\% - 1.0\%$ on the validation set, which indicates the effectiveness of each component in our proposed graph-based verification network.
We then replace the entire graph-based verification module with the concatenation of model inputs and graph node representations only to predict the verification result. This operation causes a $1.5\%$ drop on the validation set, which further shows the effectiveness of our constructed graph: i.e., catching logical relations between entities and functions plays a big part in our model.

Besides, eliminating the evidence retrieval module leads to significant drops in performance, with $4.4\%$ on the validation set. This result demonstrates that logic-level evidence, as supplementary information to the original table, plays a crucial role in our proposed approach, which can help our model better understand semi-structured tables. In addition, by removing this module, our model will be left with just the table and the statement to perform linguistic reasoning only, which also proves the importance of combining linguistic reasoning and symbolic reasoning.

\begin{table}[]
    \begin{tabular}{p{5.5cm}|c}
        \hline
        Model &  Val \\
        \hline
        LERGV & 75.6 \\
          - Node Pruning & 75.2 \\
          - Node Type Representation & 74.6 \\
          - Graph-based Verification & 74.1 \\
          - Evidence Retrieval & 71.2 \\
        \hline
    \end{tabular}
    \caption{Ablation study of the model components.}
    \label{tab:ablation}
\end{table}


\subsection{Case Study}
\begin{figure}
    \centering
    \includegraphics[width=0.45\textwidth]{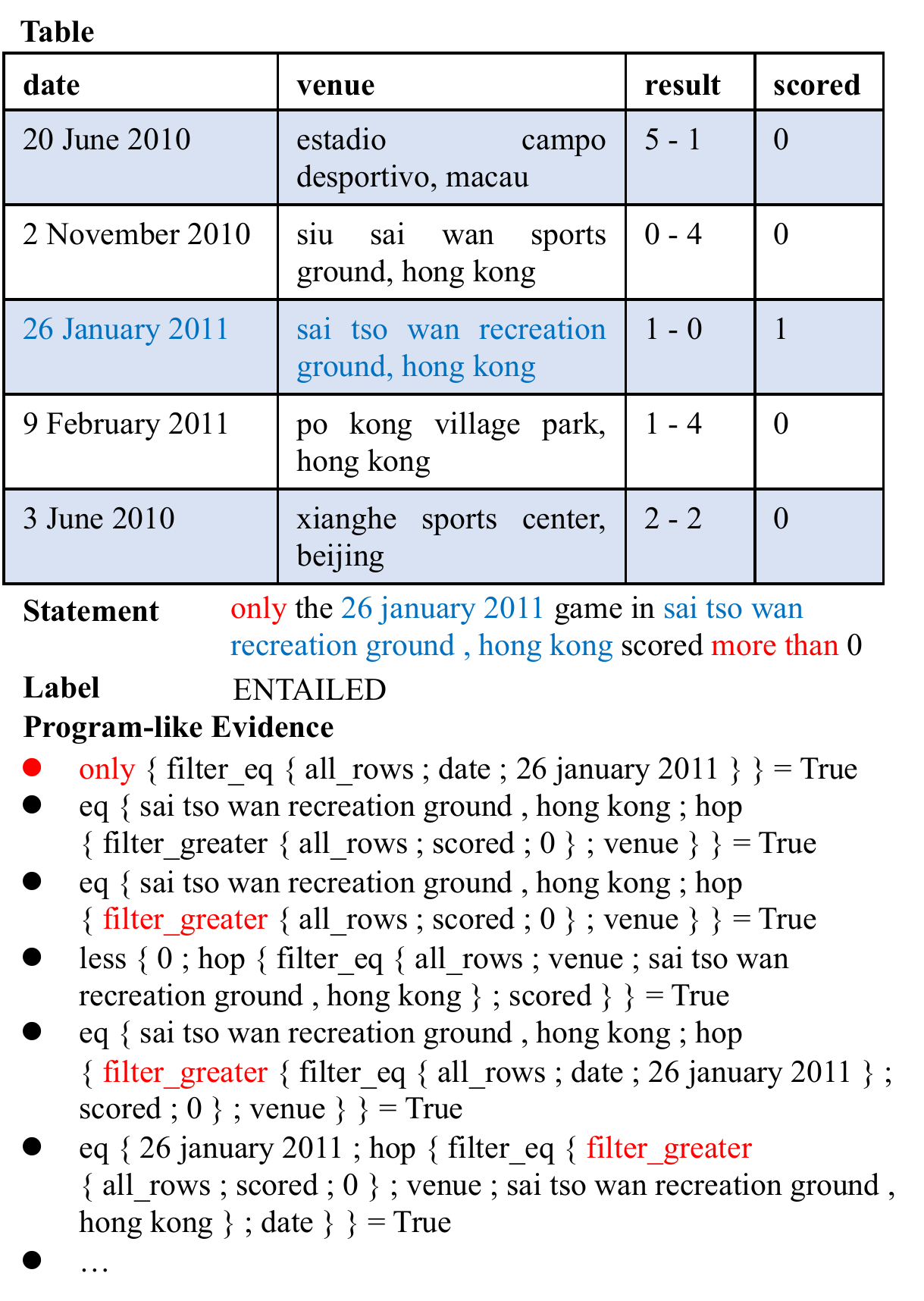}
    \caption{Case study of our proposed approach.}
    \label{case_study}
\end{figure}

We provide an example to show the quality of our proposed approach, which is shown in Figure \ref{case_study}. The key point to verify the correctness of such a statement is to obtain evidence about \emph{"only"} and \emph{"more than"}. In our retrieved evidence, the first evidence program contains the function \emph{"only"}, and can establish the logic-level connections between \emph{"26 January 2011"} and \emph{"only"} by graph construction process. Besides, three evidence programs within the function \emph{"filter\_greater"} indicate that the score of the game with the date \emph{"26 January 2011"} and the venue \emph{"sai tso wan recreation ground, hong kong"} is greater than $0$. Other evidence without the above functions describes the information in the table that is relevant to the statement, which is also helpful to perform verification.

\subsection{Error Analysis}
We randomly sample $400$ examples and categorize the errors into three classes.

In our analysis, the first category of error is the lack of evidence, or evidence is not helpful enough to verify the correctness of the given statement. This may be caused by the following two reasons. Firstly, entities in the statements are not detected and linked correctly to the table cells in the entity linking phase. Secondly, trigger words are applied to shrink the search space in the program synthesis process, which may cause some valuable information to be discarded.
For example, the statement states \emph{"ngc 1796 has the largest apparent magnitude of 12.9 followed by ngc 1705 with 12.8"}, the evidence with the function \emph{"second"} is not obtained, which causes the model to fail to get the correct prediction. $363$ error examples are caused by this category.

The second category of error appears when the statement requires numerical reasoning. For example, for the statement \emph{"jelle van damme scored three times as much as each of the other two players in the uefa champions league tournament"}, the proposed evidence retrieval approach cannot deal with the operation \emph{"three times"}. $32$ error examples are caused by this category, and we leave the numerical reasoning for future work. 

The third category of error is caused by the program denotation. In other words, the order of the arguments in the program will influence semantic understanding. For example, the statement states \emph{"A is larger than B. } The retrieved evidence is \emph{"less \{ B; A\}"}, which tends to be predicted as "REFUTED" due to the gap between \emph{"larger than"} and \emph{"less"}, although two expressions have the same meanings. $5$ error examples are caused by this category.

\section{Related Work}
\subsection{Fact Verification}
Fact verification aims to verify the correctness of the input claim by the given evidence. Most of the existing methods on fact verification focus on dealing with the unstructured text as evidence. FEVER \cite{thorne2018fever} is one of the most popular datasets in this direction, which develops automatic fact verification systems to check the veracity of claims by extracting evidence from Wikipedia. 
After that, FEVER 2.0 share task \cite{thorne2019second} is built, which is more challenging by the addition of an adversarial attack task. 
Recently, HOVER \cite{jiang2020hover} is proposed to focus on the many-hop evidence extraction and fact verification task.
Previous work mainly follows the pipeline composed of document retrieval, evidence sentence selection, and claim verification. Most of the proposed models focus on the claim verification stage and graph-based reasoning approaches \cite{zhou2019gear,liu-etal-2020-fine,zhong-etal-2020-reasoning} are popular in this stage. 

Studies on fact verification over semi-structured evidence achieve much attention recently due to the proposal of the TABFACT dataset \cite{chen2019tabfact}. Two official baselines are provided along with this dataset named Table-BERT and LPA, which treat the task in a soft linguistic reasoning manner and hard symbolic reasoning manner respectively. Some approaches on this dataset focus on the representation of the semi-structured data \cite{zhang2020table,dong2021structural}. And some approaches focus on the combination of linguistic reasoning and symbolic reasoning \cite{zhong2020logicalfactchecker,shi2020learn,yang2020program} mainly by building a semantic parser to select programs to serve the verification process. Different from their work, we propose an evidence retrieval module with a rule-based approach to obtain logic-level evidence as supplementary information for the original table to benefit the verification model.

\subsection{Reasoning over Semi-Structured Data}
Understanding semi-structured data is supposed to understand the structure and the content in every cell simultaneously. 
There are a lot of approaches in this direction spread over different tasks, such as question answering \cite{pasupat-liang-2015-compositional,nan2021fetaqa}, natural language inference \cite{gupta2020infotabs,neeraja2021incorporating}, fact verification \cite{chen2019tabfact}, etc. And some methods put attention on the pre-training strategies on the semi-structured data along with the textual input \cite{herzig2020tapas,yin2020tabert,eisenschlos2020understanding,yu2020grappa}. 
Besides, a range of approaches reason on mixed evidence sources incorporating semi-structured data, such as reasoning over table and text together \cite{chen2020hybridqa,chen2020open} and multi-modal evidence including table, text and image \cite{talmor2021multimodalqa}.
Our work treats the semi-structured table as a large evidence set and leverages the proposed evidence retrieval approach to aggregate information from tables to obtain logic-level evidence to perform verification.  




\section{Conclusion}
In this study, we focus on the table-based fact verification task and propose LERGV that leverages logic-level program-like evidence to perform fact verification over tables.
The main idea is that we formulate the task as an evidence retrieval and graph-based reasoning pipeline, and treat logic-level evidence as supplementary information for tables to avoid the issue of spurious programs. 
Specifically, we first apply a rule-based evidence retrieval approach to select, decompose, and filter among synthesized programs to obtain valuable logic-level program-like evidence.
Then, we construct a graph based on the retrieved evidence to catch logical relations between entities and functions. 
Finally, we propose a graph-based verification network to perform reasoning over the constructed graph to combine linguistic reasoning and symbolic reasoning effectively. 
Experimental results on TABFACT illustrate that our model outperforms all the baseline systems and achieves competitive results. 

\section*{Acknowledgements}
We would like to thank the anonymous reviewers for their helpful comments. This work was supported by the Key Development Program of the Ministry of Science and Technology (No. 2019YFF0303003), the National Natural Science Foundation of China (No.61976068) and "Hundreds, Millions" Engineering Science and Technology Major Special Project of Heilongjiang Province (No.2020ZX14A02).

\bibliography{anthology,custom}
\bibliographystyle{acl_natbib}




\end{document}